\documentclass[twocolumn,amsmath,nofootinbib]{revtex4} 
\usepackage{url}
\usepackage{amssymb}
\usepackage{amsfonts}
\usepackage{amsbsy}
\usepackage{graphicx}
\usepackage{hyperref}
\usepackage{array} 
\usepackage{multirow}

\newcolumntype{x}[1]{%
>{\centering\hspace{0pt}}p{#1}}%
\providecommand{\openone}{\leavevmode\hbox{\small1\kern-3.8pt\normalsize1}}

\def\spose#1{\hbox to 0pt{#1\hss}}
\def\simlt{\mathrel{\spose{\lower 3pt\hbox{$\mathchar"218$}}
   \raise 2.0pt\hbox{$\mathchar"13C$}}}
\def\simgt{\mathrel{\spose{\lower 3pt\hbox{$\mathchar"218$}}
     \raise 2.0pt\hbox{$\mathchar"13E$}}}
 \def\simpropto{\mathrel{\spose{\lower 3pt\hbox{$\mathchar"218$}}
     \raise 2.0pt\hbox{$\propto$}}}

\def\beq#1{\begin{equation}\label{#1}}
\def\eeq{\end{equation}}
\def\beqa#1{\begin{eqnarray}\label{#1}}
\def\eeqa{\end{eqnarray}}

\def\ed{\end{document}}

\def\rn{}
\def\nn#1 #2{#2. #1}				
\def\nnn#1 #2 #3{#2. #3. #1}			
\def\nnnn#1 #2 #3 #4{#2. #3. #4 #1}		
\def\nnnnn#1 #2 #3 #4 #5{#2. #3. #4 #5. #1}	

\def\rf#1;#2;#3;#4;#5 {{\frenchspacing\par\rn#1, #3 {\bf #4}, #5 (#2). \par}}
\def\rfe#1;#2;#3;#4 {{\frenchspacing\par\rn#1, #3, #4 (#2). \par}}
\def\rg#1;#2;#3;#4;#5;#6 {{\frenchspacing\par\rn#1, #3 {\bf #4}, #5 (#2). \par}}
\def\rfbook#1;#2;#3;#4;#5 {{\frenchspacing\par\rn#1, {\it #3} (#5, #4, #2).\par}}
\def\rfprep#1;#2;#3 {{\par\frenchspacing\rn#1, #3 (#2).\par}}
\def\rfproc#1;#2;#3;#4;#5;#6 {{\frenchspacing\par\rn#1 #2, in {\it #3}, ed. #4 (#5: #6)\par}}
\def\rfprocp#1;#2;#3;#4;#5;#6;#7 {{\frenchspacing\par\rn#1 #2, in {\it #3}, ed. #4 (#5: #6), p#7\par}}

\begin{document}
\pdfoptionalwaysusepdfpagebox=5


\title{Research Priorities for Robust and Beneficial Artificial Intelligence\footnote{Published in AI Magazine {\bf 36}, No 4 (2015): \url{http://tinyurl.com/rbaipaper}. This article gives examples of the type of research advocated by the Open Letter at \url{http://futureoflife.org/ai-open-letter}}
}

\author{Stuart Russell, Daniel Dewey, Max Tegmark}

\address{Computer Science Division, University of California, Berkeley, CA 94720}
\address{Dept.~of Physics \& MIT Kavli Institute, Massachusetts Institute of Technology, Cambridge, MA 02139}
\address{Future of Humanity Institute, Oxford University, 16-17 St. Ebbe's str., Oxford OX1 1PT, UK}

\begin{abstract}
Success in the quest for artificial intelligence has the potential to bring unprecedented benefits to humanity, and it is therefore worthwhile to investigate how to maximize these benefits while avoiding potential pitfalls. This article gives numerous examples (which should by no means be construed as an exhaustive list) of such worthwhile research aimed at ensuring that AI remains robust and beneficial.
\end{abstract}

\maketitle

Artificial intelligence (AI) research has explored a variety of problems and approaches since its inception, but for the last 20 years or so has been focused on the problems surrounding the construction of {\em intelligent agents} \textendash\ systems that perceive and act in some environment. In this context, the criterion for intelligence is related to statistical and economic notions of rationality \textendash\ colloquially, the ability to make good decisions, plans, or inferences.
The adoption of probabilistic representations and statistical learning methods has led to a large degree of integration and cross-fertilization between AI, machine learning, statistics, control theory, neuroscience, and other fields. The establishment of shared theoretical frameworks, combined with the availability of data and processing power, has yielded remarkable successes in various component tasks such as speech recognition, image classification, autonomous vehicles, machine translation, legged locomotion, and question-answering systems.

As capabilities in these areas and others cross the threshold from laboratory research to economically valuable technologies, a virtuous cycle takes hold whereby even small improvements in performance are worth large sums of money, prompting greater investments in research. There is now a broad consensus that AI research is progressing steadily, and that its impact on society is likely to increase. The potential benefits are huge, since everything that civilization has to offer is a product of human intelligence; we cannot predict what we might achieve when this intelligence is magnified by the tools AI may provide, but the eradication of disease and poverty are not unfathomable. Because of the great potential of AI, it is valuable to investigate how to reap its benefits while avoiding potential pitfalls.
 
Progress in AI research makes it timely to focus research not only on making AI more capable, but also on maximizing the societal benefit of AI. Such considerations motivated the AAAI 2008\textendash09 Presidential Panel on Long-Term AI Futures \citep{horvitz2009interim} and other projects and community efforts on AI's future impacts. These constitute a significant expansion of the field of AI itself, which up to now has focused largely on techniques that are neutral with respect to purpose. The present document can be viewed as a natural continuation of these efforts, focusing on identifying research directions that can help maximize the societal benefit of AI. This research is by necessity interdisciplinary, because it involves both society and AI. It ranges from economics, law, and philosophy to computer security, formal methods and, of course, various branches of AI itself. The focus is on delivering AI that is {\it beneficial} to society and {\it robust} in the sense that the benefits are guaranteed: our AI systems must do what we want them to do.

This document was drafted with input from the attendees of the 2015 conference ``The Future of AI: Opportunities and Challenges''\footnote{More details about the conference, including many of the talks, are available at \url{http://tinyurl.com/beneficialai}.} (see Acknowledgements), and was the basis for an open letter that has collected over 8,000 signatures in support of the research priorities outlined here.

\vspace{5mm}

\section{Short-term Research Priorities}

\subsection{Optimizing AI's Economic Impact}

The successes of industrial applications of AI, from manufacturing to information services, demonstrate a growing impact on the economy, although there is disagreement about the exact nature of this impact and on how to distinguish between the effects of AI and those of other information technologies. Many economists and computer scientists agree that there is valuable research to be done on how to maximize the economic benefits of AI while mitigating adverse effects, which could include increased inequality and unemployment \citep{mokyr2014secular,brynjolfsson2014second,frey2013future,glaeser2014secular,shanahan2015technological,nilsson1984artificial,manyika2013disruptive}. Such considerations motivate a range of research directions, spanning areas from economics to psychology. Below are a few examples that should by no means be interpreted as an exhaustive list.

\begin{enumerate}
\item \textbf{Labor market forecasting}: When and in what order should we expect various jobs to become automated \citep{frey2013future}? How will this affect the wages of less skilled workers, the creative professions, and different kinds of information workers? Some have have argued that AI is likely to greatly increase the overall wealth of humanity as a whole \citep{brynjolfsson2014second}. However, increased automation may push income distribution further towards a power law \citep{brynjolfsson2014labor}, and the resulting disparity may fall disproportionately along lines of race, class, and gender; research anticipating the economic and societal impact of such disparity could be useful.
\item{} \textbf{Other market disruptions}: Significant parts of the economy, including finance, insurance, actuarial, and many consumer markets, could be susceptible to disruption through the use of AI techniques to learn, model, and predict human and market behaviors. These markets might be identified by a combination of high complexity and high rewards for navigating that complexity \citep{manyika2013disruptive}.
\item \textbf{Policy for managing adverse effects}: What policies could help increasingly automated societies flourish? For example, Brynjolfsson and McAfee \citep{brynjolfsson2014second} explore various policies for incentivizing development of labor-intensive sectors and for using AI-generated wealth to support underemployed populations. What are the pros and cons of interventions such as educational reform, apprenticeship programs, labor-demanding infrastructure projects, and changes to minimum wage law, tax structure, and the social safety net \citep{glaeser2014secular}? History provides many examples of subpopulations not needing to work for economic security, ranging from aristocrats in antiquity to many present-day citizens of Qatar. What societal structures and other factors determine whether such populations flourish? Unemployment is not the same as leisure, and there are deep links between unemployment and unhappiness, self-doubt, and isolation \citep{hetschko2014changing,clark1994unhappiness}; understanding what policies and norms can break these links could significantly improve the median quality of life. Empirical and theoretical research on topics such as the basic income proposal could clarify our options \citep{van1992arguing,widerquist2013basic}.
\item \textbf{Economic measures}: It is possible that economic measures such as real GDP per capita do not accurately capture the benefits and detriments of heavily AI-and-automation-based economies, making these metrics unsuitable for policy purposes \citep{mokyr2014secular}. Research on improved metrics could be useful for decision-making.

\end{enumerate}

\subsection{Law and Ethics Research}

The development of systems that embody significant amounts of intelligence and autonomy leads to important legal and ethical questions whose answers impact both producers and consumers of AI technology. These questions span law, public policy, professional ethics, and philosophical ethics, and will require expertise from computer scientists, legal experts, political scientists, and ethicists. For example:

\begin{enumerate}
\item \textbf{Liability and law for autonomous vehicles}: If self-driving cars cut the roughly 40,000 annual US traffic fatalities in half, the car makers might get not 20,000 thank-you notes, but 20,000 lawsuits. In what legal framework can the safety benefits of autonomous vehicles such as drone aircraft and self-driving cars best be realized \citep{vladeck2014machines}? Should legal questions about AI be handled by existing (software- and internet-focused) ``cyberlaw'', or should they be treated separately \citep{calo2014robotics}? In both military and commercial applications, governments will need to decide how best to bring the relevant expertise to bear; for example, a panel or committee of professionals and academics could be created, and Calo has proposed the creation of a Federal Robotics Commission \citep{calo2014case}.
\item \textbf{Machine ethics}: How should an autonomous vehicle trade off, say, a small probability of injury to a human against the near-certainty of a large material cost? How should lawyers, ethicists, and policymakers engage the public on these issues? 
Should such trade-offs be the subject of national standards?
\item {\bf Autonomous weapons}: Can lethal autonomous weapons be made to comply with humanitarian law \citep{churchill2000autonomous}? If, as some organizations have suggested, autonomous weapons should be banned \citep{docherty2012losing}, is it possible to develop a precise definition of autonomy for this purpose, and can such a ban practically be enforced? If it is permissible or legal to use lethal autonomous weapons, how should these weapons be integrated into the existing command-and-control structure so that responsibility and liability remain associated with specific human actors? What technical realities and forecasts should inform these questions, and how should ``meaningful human control'' over weapons be defined \citep{roff2013responsibility,roff2014strategic,anderson2014adapting}? Are autonomous weapons likely to reduce political aversion to conflict, or perhaps result in ``accidental'' battles or wars \citep{asaro2008just}? Would such weapons become the tool of choice for oppressors or terrorists? Finally, how can transparency and public discourse best be encouraged on these issues?
\item {\bf Privacy}: How should the ability of AI systems to interpret the data obtained from surveillance cameras, phone lines, emails, {\it etc.}, interact with the right to privacy? How will privacy risks interact with cybersecurity and cyberwarfare \citep{singer2014cybersecurity}?
Our ability to take full advantage of the synergy between AI and big data will depend in part on our ability to manage and preserve privacy \citep{manyika2011big,agrawal2000privacy}.
\item {\bf Professional ethics}: What role should computer scientists play in the law and ethics of AI development and use? Past and current projects to explore these questions include the AAAI 2008\textendash09 Presidential Panel on Long-Term AI Futures \citep{horvitz2009interim}, the EPSRC Principles of Robotics \citep{boden2011principles}, and recently announced programs such as Stanford's One-Hundred Year Study of AI and the AAAI Committee on AI Impact and Ethical Issues.
\end{enumerate}
From a public policy perspective, AI (like any powerful new technology) enables both great new benefits and novel pitfalls to be avoided, and appropriate policies can ensure that we can enjoy the benefits while risks are minimized. This raises policy questions such as these:
\begin{enumerate}
\item What is the space of policies worth studying, and how might they be enacted?
\item Which criteria should be used to determine the merits of a policy? Candidates include
verifiability of compliance, enforceability,  ability to reduce risk, ability to avoid stifling desirable technology development, adoptability, and ability to adapt over time to changing circumstances.
\end{enumerate}

\subsection{Computer Science Research for Robust AI}

As autonomous systems become more prevalent in society, it becomes increasingly important that they robustly behave as intended. The development of autonomous vehicles, autonomous trading systems, autonomous weapons, {\it etc.} has therefore stoked interest in high-assurance systems where strong robustness guarantees can be made; Weld and Etzioni have argued that ``society will reject autonomous agents unless we have some credible means of making them safe" \citep{weld1994first}. Different ways in which an AI system may fail to perform as desired correspond to different areas of robustness research:

\begin{enumerate}
\item {\bf Verification}: how to prove that a system satisfies certain desired formal properties. ({\it ``Did I build the system right?''})
\item{\bf Validity}: how to ensure that a system that meets its formal requirements does not have unwanted behaviors and consequences. ({\it ``Did I build the right system?''})
\item{\bf Security}: how to prevent intentional manipulation by unauthorized parties.
\item{\bf Control}: how to enable meaningful human control over an AI system after it begins to operate. ({\it ``OK, I built the system wrong; can I fix it?''})
\end{enumerate}

\subsubsection{Verification}
By verification, we mean methods that yield high confidence that a system will satisfy a set of formal constraints. When possible, it is desirable for systems in safety-critical situations, e.g.\ self-driving cars, to be verifiable.

Formal verification of software has advanced significantly in recent years: examples include the {\it seL4} kernel \citep{klein2009sel4}, a complete, general-purpose operating-system kernel that has been mathematically checked against a formal specification to give a strong guarantee against crashes and unsafe operations, and HACMS, DARPA's ``clean-slate, formal methods-based approach" to a set of high-assurance software tools \citep{fisher2012hacms}. Not only should it be possible to build AI systems on top of verified substrates; it should also be possible to verify the designs of the AI systems themselves, particularly if they follow a ``componentized architecture", in which guarantees about individual components can be combined according to their connections to yield properties of the overall system. This mirrors the agent architectures used in Russell and Norvig (2010), which separate an agent into distinct modules (predictive models, state estimates, utility functions, policies, learning elements, {\it etc.}), and has analogues in some formal results on control system designs. Research on richer kinds of agents \textendash\ for example, agents with layered architectures, anytime components, overlapping deliberative and reactive elements, metalevel control, {\it etc.} \textendash\ could contribute to the creation of verifiable agents, but we lack the formal ``algebra'' to properly define, explore, and rank the space of designs.

Perhaps the most salient difference between verification of traditional software and verification of AI systems is that the correctness of traditional software is defined with respect to a fixed and known machine model, whereas AI systems \textendash\ especially robots and other embodied systems \textendash\ operate in environments that are at best partially known by the system designer. In these cases, it may be practical to verify that the system acts correctly given the knowledge that it has, avoiding the problem of modelling the real environment \citep{dennis2013practical}. A lack of design-time knowledge also motivates the use of learning algorithms within the agent software, and verification becomes more difficult: statistical learning theory gives so-called $\epsilon$-$\delta$ (probably approximately correct) bounds, mostly for the somewhat unrealistic settings of supervised learning from i.i.d.\ data and single-agent reinforcement learning with simple architectures and full observability, but even then requiring prohibitively large sample sizes to obtain meaningful guarantees.

Work in adaptive control theory \citep{aastrom2013adaptive}, the theory of so-called {\em cyberphysical systems} \citep{platzer2010logical}, and verification of hybrid or robotic systems \citep{alur2011formal, winfield2014towards} is highly relevant but also faces the same difficulties. And of course all these issues are laid on top of the standard problem of proving that a given software artifact does in fact correctly implement, say, a reinforcement learning algorithm of the intended type. Some work has been done on verifying neural network applications \citep{pulina2010abstraction, taylor2006methods, schumann2010applications} and the notion of {\em partial programs}~\citep{andre2002state,spears2006assuring} allows the designer to impose arbitrary ``structural'' constraints on behavior, but much remains to be done before it will be possible to have high confidence that a learning agent will learn to satisfy its design criteria in realistic contexts. 

\subsubsection{Validity}

A verification theorem for an agent design has the form, ``If
environment satisfies assumptions $\phi$ then behavior satisfies
requirements $\psi$.'' There are two ways in which a verified agent
can, nonetheless, fail to be a beneficial agent in actuality: first,
the environmental assumption $\phi$ is false in the real world,
leading to behavior that violates the requirements $\psi$; second, the
system may satisfy the formal requirement $\psi$ but still behave in
ways that we find highly undesirable in practice. 
It may be the case that
this undesirability is a consequence of satisfying $\psi$ when
$\phi$ is violated; i.e., had $\phi$ held the undesirability would not have been manifested; or it may be the case
that the requirement $\psi$ is erroneous in itself. Russell and Norvig (2010) provide a simple example: if a robot vacuum cleaner is
asked to clean up as much dirt as possible, and has an action to dump the contents of its dirt container, it will repeatedly dump and clean up the same dirt.
The requirement should focus not on dirt cleaned up but on cleanliness of the floor.
Such specification errors are ubiquitous in software verification, where it is commonly observed that writing correct specifications can be
harder than writing correct code. Unfortunately, it is not possible to verify the specification:
the notions of ``beneficial'' and ``desirable'' are not separately made formal,
so one cannot straightforwardly prove that satisfying $\psi$ necessarily leads to desirable behavior and a beneficial agent.


In order to build systems that robustly behave well, we of course need to decide what ``good behavior'' means in each application domain. This ethical question is tied intimately to questions of what engineering techniques are available, how reliable these techniques are, and what trade-offs can be made \textendash\ all areas where computer science, machine learning, and broader AI expertise is valuable. For example, Wallach and Allen (2008) argue that a significant consideration is the computational expense of different behavioral standards (or ethical theories): if a standard cannot be applied efficiently enough to guide behavior in safety-critical situations, then cheaper approximations may be needed. Designing simplified rules \textendash\ for example, to govern a self-driving car's decisions in critical situations \textendash\ will likely require expertise from both ethicists and computer scientists. Computational models of ethical reasoning may shed light on questions of computational expense and the viability of reliable ethical reasoning methods \citep{asaro2006should, sullins2011introduction}.


\subsubsection{Security}

Security research can help make AI more robust.
As AI systems are used in an increasing number of critical roles, they will take up an increasing proportion of cyber-attack surface area. It is also probable that AI and machine learning techniques will themselves be used in cyber-attacks. 

Robustness against exploitation at the low level is closely tied to verifiability and freedom from bugs. For example, the DARPA SAFE program aims to build an integrated hardware-software system with a flexible metadata rule engine, on which can be built memory safety, fault isolation, and other protocols that could improve security by preventing exploitable flaws \citep{dehon2011preliminary}. Such programs cannot eliminate all security flaws (since verification is only as strong as the assumptions that underly the specification), but could significantly reduce vulnerabilities of the type exploited by the recent ``Heartbleed'' and ``Bash'' bugs. Such systems could be preferentially deployed in safety-critical applications, where the cost of improved security is justified.

At a higher level, research into specific AI and machine learning techniques may become increasingly useful in security. These techniques could be applied to the detection of intrusions \citep{lane2000machine}, analyzing malware \citep{rieck2011automatic}, or detecting potential exploits in other programs through code analysis \citep{brun2004finding}. It is not implausible that cyberattack between states and private actors will be a risk factor for harm from near-future AI systems, motivating research on preventing harmful events. As AI systems grow more complex and are networked together, they will have to intelligently manage their trust, motivating research on statistical-behavioral trust establishment  \citep{probst2007statistical} and computational reputation models \citep{sabater2005review}.

\subsubsection{Control}

For certain types of safety-critical AI systems \textendash\ especially vehicles and weapons platforms \textendash\ it may be desirable to retain some form of meaningful human control, whether this means a human in the loop, on the loop \citep{hexmoor2009natural,parasuraman2000model}, or some other protocol. In any of these cases, there will be technical work needed in order to ensure that meaningful human control is maintained \citep{united2014weaponization}.
 
Automated vehicles are a test-bed for effective control-granting techniques. The design of systems and protocols for transition between automated navigation and human control is a promising area for further research. Such issues also motivate broader research on how to optimally allocate tasks within human\textendash computer teams, both for identifying situations where control should be transferred, and for applying human judgment efficiently to the highest-value decisions.

\section{Long-term research priorities}

A frequently discussed long-term goal of some AI researchers is to develop systems that can learn from experience with human-like breadth and surpass human performance in most cognitive tasks, thereby having a major impact on society.
If there is a non-negligible probability that these efforts will succeed in the foreseeable future, then additional current research beyond that mentioned in the previous sections will be motivated as exemplified below, to help ensure that the resulting AI will be robust and beneficial.

Assessments of this success probability vary widely between researchers, but few would argue with great confidence that the probability is negligible, given the track record of such predictions. For example, Ernest Rutherford, arguably the greatest nuclear physicist of his time, said in 1933 \textendash\ less than 24 hours before Szilard's invention of the nuclear chain reaction \textendash\ that nuclear energy was ``moonshine'' \citep{ap1933atom}, and
Astronomer Royal Richard Woolley called interplanetary travel ``utter bilge'' in 1956 \citep{Woolley1956}.
Moreover, to justify a modest investment in this AI robustness research, this probability need not be high, merely non-negligible, just as a modest investment in home insurance is justified by a non-negligible probability of the home burning down.

\subsection{Verification}

Reprising the themes of short-term research, research enabling verifiable low-level software and hardware can eliminate large classes of bugs and problems in general AI systems; if such systems become increasingly powerful and safety-critical, verifiable safety properties will become increasingly valuable. If the theory of extending verifiable properties from components to entire systems is well understood, then even very large systems can enjoy certain kinds of safety guarantees, potentially aided by techniques designed explicitly to handle learning agents and high-level properties. Theoretical research, especially if it is done explicitly with very general and capable AI systems in mind, could be particularly useful.

A related verification research topic that is distinctive to long-term concerns is the verifiability of systems that modify, extend, or improve themselves, possibly many times in succession \citep{Good1965,Vinge1993}. Attempting to straightforwardly apply formal verification tools to this more general setting presents new difficulties, including the challenge that a formal system that is sufficiently powerful cannot use formal methods in the obvious way to gain assurance about the accuracy of functionally similar formal systems, on pain of 
inconsistency via G\"odel's incompleteness \citep{fallenstein2014vingean,weaver2013paradoxes}. It is not yet clear whether or how this problem can be overcome, or whether similar problems will arise with other verification methods of similar strength.

Finally, it is often difficult to actually apply formal verification techniques to physical systems, especially systems that have not been designed with verification in mind. This motivates research pursuing a general theory that links functional specification to physical states of affairs. This type of theory would allow use of formal tools to anticipate and control behaviors of systems that approximate rational agents, alternate designs such as satisficing agents, and systems that cannot be easily described in the standard agent formalism (powerful prediction systems, theorem-provers, limited-purpose science or engineering systems, {\it etc.}). It may also be that such a theory could allow rigorous demonstrations that systems are constrained from taking certain kinds of actions or performing certain kinds of reasoning.

\subsection{Validity}

As in the short-term research priorities, validity is concerned with undesirable behaviors that can arise despite a system's formal correctness. In the long term, AI systems might become more powerful and autonomous, in which case failures of validity could carry correspondingly higher costs.

Strong guarantees for machine learning methods, an area we highlighted for short-term validity research, will also be important for long-term safety. To maximize the long-term value of this work, machine learning research might focus on the types of unexpected generalization that would be most problematic for very general and capable AI systems. In particular, it might aim to understand theoretically and practically how learned representations of high-level human concepts could be expected to generalize (or fail to) in radically new contexts \citep{tegmark2015friendly}. Additionally, if some concepts could be learned reliably, it might be possible to use them to define tasks and constraints that minimize the chances of unintended consequences even when autonomous AI systems become very general and capable. Little work has been done on this topic, which suggests that both theoretical and experimental research may be useful.

Mathematical tools such as formal logic, probability, and decision theory have yielded significant insight into the foundations of reasoning and decision-making. However, there are still many open problems in the foundations of reasoning and decision. Solutions to these problems may make the behavior of very capable systems much more reliable and predictable. Example research topics in this area include reasoning and decision under bounded computational resources \`a la Horvitz and Russell \citep{horvitz1987reasoning,russell1995provably}, how to take into account correlations between AI systems' behaviors and those of their environments or of other agents \citep{tennenholtz2004program,AAAIW148833,hintze2014problem,halpern2013game,soares2014toward}, how agents that are embedded in their environments should reason \citep{soares2014formalizing,orseau2012space}, and how to reason about uncertainty over logical consequences of beliefs or other deterministic computations \citep{soares2014questions}. These topics may benefit from being considered together, since they appear deeply linked \citep{halpern2011don, halpern2014decision}.

In the long term, it is plausible that we will want to make agents that act autonomously and powerfully across many domains. Explicitly specifying our preferences in broad domains in the style of near-future machine ethics may not be practical, making ``aligning'' the values of powerful AI systems with our own values and preferences difficult \citep{soares2014value, soares2014aligning}. Consider, for instance, the difficulty of creating a utility function that encompasses an entire body of law; even a literal rendition of the law is far beyond our current capabilities, and would be highly unsatisfactory in practice (since law is written assuming that it will be interpreted and applied in a flexible, case-by-case way by humans who, presumably, already embody the background value systems that artificial agents may lack). Reinforcement learning raises its own problems: when systems become very capable and general, then an effect similar to Goodhart's Law is likely to occur, in which sophisticated agents attempt to manipulate or directly control their reward signals \citep{bostrom2014superintelligence}. This motivates research areas that could improve our ability to engineer systems that can learn or acquire values at run-time. For example, inverse reinforcement learning may offer a viable approach, in which a system infers the preferences of another rational or nearly rational actor by observing its behavior \citep{russell1998learning,ng2000algorithms}. Other approaches could use different assumptions about underlying cognitive models of the actor whose preferences are being learned \citep{chu2005preference}, or could be explicitly inspired by the way humans acquire ethical values. As systems become more capable, more epistemically difficult methods could become viable, suggesting that research on such methods could be useful; for example, Bostrom (2014) reviews preliminary work on a variety of methods for specifying goals indirectly.

\subsection{Security}
It is unclear whether long-term progress in AI will make the overall problem of security easier or harder; on one hand, systems will become increasingly complex in construction and behavior and AI-based cyberattacks may be extremely effective, while on the other hand, the use of AI and machine learning techniques along with significant progress in low-level system reliability may render hardened systems much less vulnerable than today's. From a cryptographic perspective, it appears that this conflict favors defenders over attackers; this may be a reason to pursue effective defense research wholeheartedly.

Although the topics described in the near-term security research section above may become increasingly important in the long term, very general and capable systems will pose distinctive security problems. In particular, if the problems of validity and control are not solved, it may be useful to create ``containers" for AI systems that could have undesirable behaviors and consequences in less controlled environments \citep{yampolskiy2012leakproofing}. Both theoretical and practical sides of this question warrant investigation. If the general case of AI containment turns out to be prohibitively difficult, then it may be that designing an AI system and a container in parallel is more successful, allowing the weaknesses and strengths of the design to inform the containment strategy \citep{bostrom2014superintelligence}. The design of anomaly detection systems and automated exploit-checkers could be of significant help. Overall, it seems reasonable to expect this additional perspective \textendash\ defending against attacks from ``within" a system as well as from external actors \textendash\ will raise interesting and profitable questions in the field of computer security.

\subsection{Control}

It has been argued that very general and capable AI systems operating autonomously to accomplish some task will often be subject to effects that increase the difficulty of maintaining meaningful human control \citep{omohundro2007nature,bostrom2012superintelligent,bostrom2014superintelligence,shanahan2015technological}. Research on systems that are not subject to these effects, minimize their impact, or allow for reliable human control could be valuable in preventing undesired consequences, as could work on reliable and secure test-beds for AI systems at a variety of capability levels.

If an AI system is selecting the actions that best allow it to complete a given task, then avoiding conditions that prevent the system from continuing to pursue the task is a natural subgoal \citep{omohundro2007nature,bostrom2012superintelligent} (and conversely, seeking unconstrained situations is sometimes a useful heuristic \citep{wissner2013causal}). This could become problematic, however, if we wish to repurpose the system, to deactivate it, or to significantly alter its decision-making process; such a system would rationally avoid these changes. Systems that do not exhibit these behaviors have been termed {\it corrigible} systems \citep{soares2014corrigibility}, and both theoretical and practical work in this area appears tractable and useful. For example, it may be possible to design utility functions or decision processes so that a system will not try to avoid being shut down or repurposed \citep{soares2014corrigibility}, and theoretical frameworks could be developed to better understand the space of potential systems that avoid undesirable behaviors \citep{hibbard2012avoiding,hibbard2014ethical,hibbard2015self}.

It has been argued that another natural subgoal for AI systems pursuing a given goal is the acquisition of fungible resources of a variety of kinds: for example, information about the environment, safety from disruption, and improved freedom of action are all instrumentally useful for many tasks \citep{omohundro2007nature,bostrom2012superintelligent}. Hammond et al (1995) gives the label {\it stabilization} to the more general set of cases where ``due to the action of the agent, the environment comes to be better fitted to the agent as time goes on''. This type of subgoal could lead to undesired consequences, and a better understanding of the conditions under which resource acquisition or radical stabilization is an optimal strategy (or likely to be selected by a given system) would be useful in mitigating its effects. Potential research topics in this area include ``domestic" goals that are limited in scope in some way \citep{bostrom2014superintelligence}, the effects of large temporal discount rates on resource acquisition strategies, and experimental investigation of simple systems that display these subgoals.

Finally, research on the possibility of superintelligent machines or rapid, sustained self-improvement (``intelligence explosion'') has been highlighted by past and current projects on the future of AI as potentially valuable to the project of maintaining reliable control in the long term. The AAAI 2008\textendash09 Presidential Panel on Long-Term AI Futures' ``Subgroup on Pace, Concerns, and Control'' stated that\begin{quote}There was overall skepticism about the prospect of an intelligence explosion... Nevertheless, there was a shared sense that additional research would be valuable on methods for understanding and verifying the range of behaviors of complex computational systems to minimize unexpected outcomes. Some panelists recommended that more research needs to be done to better define ``intelligence explosion,'' and also to better formulate different classes of such accelerating intelligences. Technical work would likely lead to enhanced understanding of the likelihood of such phenomena, and the nature, risks, and overall outcomes associated with different conceived variants  \citep{horvitz2009interim}.
\end{quote}Stanford's One-Hundred Year Study of Artificial Intelligence includes ``Loss of Control of AI systems'' as an area of study, specifically highlighting concerns over the possibility that
\begin{quote}...we could one day lose control of AI systems via the rise of superintelligences that do not act in accordance with human wishes \textendash\ and that such powerful systems would threaten humanity. Are such dystopic outcomes possible?  If so, how might these situations arise? ...What kind of investments in research should be made to better understand and to address the possibility of the rise of a dangerous superintelligence or the occurrence of an ``intelligence explosion''? \citep{horvitz2014hundred}\end{quote}
Research in this area could include any of the long-term research priorities listed above, as well as theoretical and forecasting work on intelligence explosion and superintelligence \citep{chalmers2010singularity, bostrom2014superintelligence}, and could extend or critique existing approaches begun by groups such as the Machine Intelligence Research Institute \citep{soares2014aligning}.

\section{Conclusion}


In summary, success in the quest for artificial intelligence has the potential to bring unprecedented benefits to humanity, and it is therefore worthwhile to research how to maximize these benefits while avoiding potential pitfalls. The research agenda outlined in this paper, and the concerns that motivate it, have been called ``anti-AI", but we vigorously contest this characterization. It seems self-evident that the growing capabilities of AI are leading to an increased potential for impact on human society. It is the duty of AI researchers to ensure that the future impact is beneficial. We believe that this is possible, and hope that this research agenda provides a helpful step in the right direction.


\section{Authors}

{\bf Stuart Russell} is a Professor of Computer Science at UC Berkeley.  His research covers many aspects of artificial intelligence and machine learning. He is a fellow of AAAI, ACM, and AAAS and winner of the IJCAI Computers and Thought Award. He held the Chaire Blaise Pascal in Paris from 2012 to 2014. His book {\it Artificial Intelligence: A Modern Approach} (with Peter Norvig) is the standard text in the field.

{\bf Daniel Dewey} is the Alexander Tamas Research Fellow on Machine Superintelligence and the Future of AI at Oxford's Future of Humanity Institute, Oxford Martin School. He was previously at Google, Intel Labs Pittsburgh, and Carnegie Mellon University.

{\bf Max Tegmark} is a professor of physics at MIT. His current research is at the interface of physics and artificial intelligence, using physics-based techniques to explore connections between information processing in biological and engineered systems. He is the president of the Future of Life Institute, which supports research advancing robust and beneficial artificial intelligence.

\section{Acknowledgements}

The initial version of this document was drafted with major input from Janos Kramar and Richard Mallah, and reflects valuable feedback from Anthony Aguirre, Erik Brynjolfsson, Ryan Calo, Meia Chita-Tegmark, Tom Dietterich, Dileep George, Bill Hibbard, Demis Hassabis, Eric Horvitz, Leslie Pack Kaelbling, James Manyika, Luke Muehlhauser, Michael Osborne, David Parkes, Heather Roff, Francesca Rossi, Bart Selman, Murray Shanahan, and many others. The authors are also grateful to  Serkan Cabi and David Stanley for help with manuscript editing and formatting.

\bibliography{priorities}

\begin{thebibliography}{10}

\bibitem{horvitz2009interim}
{\sc E.~Horvitz} and {\sc B.~Selman},
\newblock Interim Report from the Panel Chairs, 2009,
\newblock AAAI Presidential Panel on Long Term AI Futures.

\bibitem{mokyr2014secular}
{\sc J.~Mokyr},
\newblock {\em Secular Stagnation: Facts, Causes and Cures} , 83 (2014).

\bibitem{brynjolfsson2014second}
{\sc E.~Brynjolfsson} and {\sc A.~McAfee},
\newblock {\em The second machine age: work, progress, and prosperity in a time
  of brilliant technologies},
\newblock W.W. Norton \& Company, 2014.

\bibitem{frey2013future}
{\sc C.~Frey} and {\sc M.~Osborne},
\newblock The future of employment: how susceptible are jobs to
  computerisation?,
\newblock Technical report, Oxford Martin School, University of Oxford, 2013.

\bibitem{glaeser2014secular}
{\sc E.~L. Glaeser},
\newblock {\em Secular Stagnation: Facts, Causes and Cures} , 69 (2014).

\bibitem{shanahan2015technological}
{\sc M.~Shanahan},
\newblock {\em The Technological Singularity},
\newblock MIT Press, 2015,
\newblock Forthcoming.

\bibitem{nilsson1984artificial}
{\sc N.~J. Nilsson},
\newblock {\em AI Magazine} {\bf 5}, 5 (1984).

\bibitem{manyika2013disruptive}
{\sc J.~Manyika}, {\sc M.~Chui}, {\sc J.~Bughin}, {\sc R.~Dobbs}, {\sc
  P.~Bisson}, and {\sc A.~Marrs},
\newblock {\em Disruptive Technologies: Advances that will Transform Life,
  Business, and the Global Economy},
\newblock McKinsey Global Institute, Washington, D.C., 2013.

\bibitem{brynjolfsson2014labor}
{\sc E.~Brynjolfsson}, {\sc A.~McAfee}, and {\sc M.~Spence},
\newblock {\em Foreign Aff.} {\bf 93}, 44 (2014).

\bibitem{hetschko2014changing}
{\sc C.~Hetschko}, {\sc A.~Knabe}, and {\sc R.~Sch{\"o}b},
\newblock {\em The Economic Journal} {\bf 124}, 149\textendash166 (2014).

\bibitem{clark1994unhappiness}
{\sc A.~E. Clark} and {\sc A.~J. Oswald},
\newblock {\em The Economic Journal} , 648\textendash659 (1994).

\bibitem{van1992arguing}
{\sc P.~Van~Parijs},
\newblock {\em Arguing for Basic Income. Ethical foundations for a radical
  reform},
\newblock Verso, 1992.

\bibitem{widerquist2013basic}
{\sc K.~Widerquist}, {\sc J.~A. Noguera}, {\sc Y.~Vanderborght}, and {\sc
  J.~De~Wispelaere},
\newblock {\em Basic income: an anthology of contemporary research},
\newblock Wiley/Blackwell, 2013.

\bibitem{vladeck2014machines}
{\sc D.~C. Vladeck},
\newblock {\em Wash. L. Rev.} {\bf 89}, 117 (2014).

\bibitem{calo2014robotics}
{\sc R.~Calo},
\newblock {\em Available at SSRN 2402972}  (2014).

\bibitem{calo2014case}
{\sc R.~Calo},
\newblock {\em Available at SSRN 2529151}  (2014).

\bibitem{churchill2000autonomous}
{\sc R.~R. Churchill} and {\sc G.~Ulfstein},
\newblock {\em American Journal of International Law} {\bf 94},
  623\textendash659 (2000).

\bibitem{docherty2012losing}
{\sc B.~L. Docherty},
\newblock {\em Losing Humanity: The Case Against Killer Robots},
\newblock Human Rights Watch, New York, 2012.

\bibitem{roff2013responsibility}
{\sc H.~M. Roff},
\newblock {\em Routledge Handbook of Ethics and War: Just War Theory in the
  21st Century} , 352 (2013).

\bibitem{roff2014strategic}
{\sc H.~M. Roff},
\newblock {\em Journal of Military Ethics} {\bf 13} (2014).

\bibitem{anderson2014adapting}
{\sc K.~Anderson}, {\sc D.~Reisner}, and {\sc M.~C. Waxman},
\newblock {\em International Law Studies} {\bf 90}, 386\textendash411 (2014).

\bibitem{asaro2008just}
{\sc P.~Asaro},
\newblock How Just could a Robot War Be?,
\newblock in {\em Current Issues in Computing And Philosophy}, edited by {\sc
  K.~W. Adam~Briggle} and {\sc P.~A.~E. Brey}, p. 50\textendash64, IOS Press,
  Amsterdam, 2008.

\bibitem{singer2014cybersecurity}
{\sc P.~W. Singer} and {\sc A.~Friedman},
\newblock {\em Cybersecurity: What Everyone Needs to Know},
\newblock Oxford University Press, New York, 2014.

\bibitem{manyika2011big}
{\sc J.~Manyika}, {\sc M.~Chui}, {\sc B.~Brown}, {\sc J.~Bughin}, {\sc
  R.~Dobbs}, {\sc C.~Roxburgh}, and {\sc A.~H. Byers},
\newblock Big Data: The Next Frontier for Innovation, Competition, and
  Productivity,
\newblock Report, McKinsey Global Institute, Washington, D.C., 2011.

\bibitem{agrawal2000privacy}
{\sc R.~Agrawal} and {\sc R.~Srikant},
\newblock {\em ACM Sigmod Record} {\bf 29}, 439\textendash450 (2000).

\bibitem{boden2011principles}
{\sc M.~Boden}, {\sc J.~Bryson}, {\sc D.~Caldwell}, {\sc K.~Dautenhahn}, {\sc
  L.~Edwards}, {\sc S.~Kember}, {\sc P.~Newman}, {\sc V.~Parry}, {\sc
  G.~Pegman}, {\sc T.~Rodden}, et~al.,
\newblock (2011).

\bibitem{weld1994first}
{\sc D.~Weld} and {\sc O.~Etzioni},
\newblock {\em AAAI Technical Report} {\bf SS-94-03}, 17\textendash23 (1994).

\bibitem{klein2009sel4}
{\sc G.~Klein}, {\sc K.~Elphinstone}, {\sc G.~Heiser}, {\sc J.~Andronick}, {\sc
  D.~Cock}, {\sc P.~Derrin}, {\sc D.~Elkaduwe}, {\sc K.~Engelhardt}, {\sc
  R.~Kolanski}, {\sc M.~Norrish}, {\sc T.~Sewell}, {\sc H.~Tuch}, and {\sc
  S.~Winwood},
\newblock seL4: Formal verification of an OS kernel,
\newblock in {\em Proceedings of the ACM SIGOPS 22nd symposium on Operating
  systems principles}, p. 207\textendash220, ACM, 2009.

\bibitem{fisher2012hacms}
{\sc K.~Fisher},
\newblock HACMS: High Assurance Cyber Military Systems,
\newblock in {\em Proceedings of the 2012 ACM Conference on High Integrity
  Language Technology}, p. 51\textendash52, ACM, 2012.

\bibitem{dennis2013practical}
{\sc L.~A. Dennis}, {\sc M.~Fisher}, {\sc N.~K. Lincoln}, {\sc A.~Lisitsa}, and
  {\sc S.~M. Veres},
\newblock {\em arXiv preprint arXiv:1310.2431}  (2013).

\bibitem{aastrom2013adaptive}
{\sc K.~J. {\AA}str{\"o}m} and {\sc B.~Wittenmark},
\newblock {\em Adaptive control},
\newblock Courier Dover Publications, 2013.

\bibitem{platzer2010logical}
{\sc A.~Platzer},
\newblock {\em Logical Analysis of Hybrid Systems: Proving Theorems for Complex
  Dynamics},
\newblock Springer, 2010.

\bibitem{alur2011formal}
{\sc R.~Alur},
\newblock Formal verification of hybrid systems,
\newblock in {\em Embedded Software (EMSOFT), 2011 Proceedings of the
  International Conference on}, p. 273\textendash278, IEEE, 2011.

\bibitem{winfield2014towards}
{\sc A.~F. Winfield}, {\sc C.~Blum}, and {\sc W.~Liu},
\newblock Towards an Ethical Robot: Internal Models, Consequences and Ethical
  Action Selection,
\newblock in {\em Advances in Autonomous Robotics Systems}, edited by {\sc
  M.~Mistry}, {\sc A.~Leonardis}, {\sc M.~Witkowski}, and {\sc C.~Melhuish}, p.
  85\textendash96, Springer, 2014.

\bibitem{pulina2010abstraction}
{\sc L.~Pulina} and {\sc A.~Tacchella},
\newblock An abstraction-refinement approach to verification of artificial
  neural networks,
\newblock in {\em Computer Aided Verification}, p. 243\textendash257, 2010.

\bibitem{taylor2006methods}
{\sc B.~J.~E. Taylor},
\newblock {\em Methods and Procedures for the Verification and Validation of
  Artificial Neural Networks},
\newblock Springer, 2006.

\bibitem{schumann2010applications}
{\sc J.~M. Schumann} and {\sc Y.~Liu},
\newblock {\em Applications of neural networks in high assurance systems},
\newblock Springer, 2010.

\bibitem{andre2002state}
{\sc D.~Andre} and {\sc S.~J. Russell},
\newblock State abstraction for programmable reinforcement learning agents,
\newblock in {\em Eighteenth national conference on Artificial intelligence},
  p. 119\textendash125, American Association for Artificial Intelligence, 2002.

\bibitem{spears2006assuring}
{\sc D.~F. Spears},
\newblock Assuring the Behavior of Adaptive Agents,
\newblock in {\em Agent Technology from a Formal Perspective}, p.
  227\textendash257, Springer, 2006.

\bibitem{asaro2006should}
{\sc P.~M. Asaro},
\newblock {\em International Review of Information Ethics} {\bf 6},
  9\textendash16 (2006).

\bibitem{sullins2011introduction}
{\sc J.~P. Sullins},
\newblock {\em Philosophy \& Technology} {\bf 24}, 233\textendash238 (2011).

\bibitem{dehon2011preliminary}
{\sc A.~DeHon}, {\sc B.~Karel}, {\sc T.~F. Knight~Jr}, {\sc G.~Malecha}, {\sc
  B.~Montagu}, {\sc R.~Morisset}, {\sc G.~Morrisett}, {\sc B.~C. Pierce}, {\sc
  R.~Pollack}, {\sc S.~Ray}, {\sc O.~Shivers}, and {\sc J.~M. Smith},
\newblock Preliminary Design of the SAFE Platform,
\newblock in {\em Proceedings of the 6th Workshop on Programming Languages and
  Operating Systems}, ACM, 2011.

\bibitem{lane2000machine}
{\sc T.~D. Lane},
\newblock Machine learning techniques for the computer security domain of
  anomaly detection, 2000,
\newblock Ph.D. Dissertation, Department of Electrical Engineering, Purdue
  University.

\bibitem{rieck2011automatic}
{\sc K.~Rieck}, {\sc P.~Trinius}, {\sc C.~Willems}, and {\sc T.~Holz},
\newblock {\em Journal of Computer Security} {\bf 19}, 639\textendash668
  (2011).

\bibitem{brun2004finding}
{\sc Y.~Brun} and {\sc M.~D. Ernst},
\newblock Finding Latent Code Errors via Machine Learning Over Program
  Executions,
\newblock in {\em Proceedings of the 26th International Conference on Software
  Engineering}, p. 480\textendash490, IEEE Computer Society, 2004.

\bibitem{probst2007statistical}
{\sc M.~J. Probst} and {\sc S.~K. Kasera},
\newblock Statistical trust establishment in wireless sensor networks,
\newblock in {\em Parallel and Distributed Systems, 2007 International
  Conference on}, volume~2, p. 1\textendash8, IEEE, 2007.

\bibitem{sabater2005review}
{\sc J.~Sabater} and {\sc C.~Sierra},
\newblock {\em Artificial intelligence review} {\bf 24}, 33\textendash60
  (2005).

\bibitem{hexmoor2009natural}
{\sc H.~Hexmoor}, {\sc B.~McLaughlan}, and {\sc G.~Tuli},
\newblock {\em Journal of Experimental \& Theoretical Artificial Intelligence}
  {\bf 21}, 59\textendash77 (2009).

\bibitem{parasuraman2000model}
{\sc R.~Parasuraman}, {\sc T.~B. Sheridan}, and {\sc C.~D. Wickens},
\newblock {\em Systems, Man and Cybernetics, Part A: Systems and Humans, IEEE
  Transactions on} {\bf 30}, 286\textendash297 (2000).

\bibitem{united2014weaponization}
{\sc UNIDIR},
\newblock {\em The Weaponization of Increasingly Autonomous Technologies:
  Implications for Security and Arms Control},
\newblock UNIDIR, 2014.

\bibitem{ap1933atom}
{\sc A.~Press},
\newblock {\em New York Herald Tribune}  (1933),
\newblock September 12, p. 1.

\bibitem{Woolley1956}
{\sc Reuters},
\newblock {\em The Ottawa Citizen}  (1956),
\newblock January 3, p. 1.

\bibitem{Good1965}
{\sc I.~J. Good},
\newblock {\em Advances in Computers} {\bf 6}, 31\textendash88 (1965).

\bibitem{Vinge1993}
{\sc V.~Vinge},
\newblock The Coming Technological Singularity,
\newblock in {\em VISION-21 Symposium, NASA Lewis Research Center and the Ohio
  Aerospace Institute}, 1993,
\newblock NASA CP-10129.

\bibitem{fallenstein2014vingean}
{\sc B.~Fallenstein} and {\sc N.~Soares},
\newblock Vingean Reflection: Reliable Reasoning for Self-Modifying Agents,
\newblock Technical report, Machine Intelligence Research Institute, Berkeley,
  2014.

\bibitem{weaver2013paradoxes}
{\sc N.~Weaver},
\newblock Paradoxes of rational agency and formal systems that verify their own
  soundness, 2013,
\newblock Preprint.

\bibitem{tegmark2015friendly}
{\sc M.~Tegmark},
\newblock Friendly Artificial Intelligence: the Physics Challenge,
\newblock in {\em Proceedings of the AAAI-15 Workshop on AI and Ethics}, p.
  87\textendash89, AAAI, 2015.

\bibitem{horvitz1987reasoning}
{\sc E.~J. Horvitz},
\newblock Reasoning About Beliefs and Actions Under Computational Resource
  Constraints,
\newblock in {\em Third AAAI Workshop on Uncertainty in Artificial
  Intelligence}, p. 429\textendash444, 1987.

\bibitem{russell1995provably}
{\sc S.~J. Russell} and {\sc D.~Subramanian},
\newblock {\em Journal of Artificial Intelligence Research} , 1\textendash36
  (1995).

\bibitem{tennenholtz2004program}
{\sc M.~Tennenholtz},
\newblock {\em Games and Economic Behavior} {\bf 49}, 363\textendash373 (2004).

\bibitem{AAAIW148833}
{\sc P.~LaVictoire}, {\sc B.~Fallenstein}, {\sc E.~Yudkowsky}, {\sc M.~Barasz},
  {\sc P.~Christiano}, and {\sc M.~Herreshoff},
\newblock Program Equilibrium in the Prisoner's Dilemma via Löb's Theorem,
\newblock in {\em AAAI Multiagent Interaction without Prior Coordination
  workshop}, 2014.

\bibitem{hintze2014problem}
{\sc D.~Hintze},
\newblock Problem Class Dominance in Predictive Dilemmas, 2014,
\newblock Honors Thesis, Arizona State University.

\bibitem{halpern2013game}
{\sc J.~Y. Halpern} and {\sc R.~Pass},
\newblock {\em arXiv preprint arXiv:1308.3778}  (2013).

\bibitem{soares2014toward}
{\sc N.~Soares} and {\sc B.~Fallenstein},
\newblock Toward Idealized Decision Theory,
\newblock Technical report, Machine Intelligence Research Institute, Berkeley,
  2014.

\bibitem{soares2014formalizing}
{\sc N.~Soares},
\newblock Formalizing Two Problems of Realistic World-Models,
\newblock Technical report, Machine Intelligence Research Institute, Berkeley,
  2014.

\bibitem{orseau2012space}
{\sc L.~Orseau} and {\sc M.~Ring},
\newblock Space-Time Embedded Intelligence,
\newblock in {\em Proceedings of the 5th International Conference on Artificial
  General Intelligence}, p. 209\textendash218, Berlin, 2012, Springer.

\bibitem{soares2014questions}
{\sc N.~Soares} and {\sc B.~Fallenstein},
\newblock Questions of Reasoning Under Logical Uncertainty,
\newblock Technical report, Machine Intelligence Research Institute, 2014,
\newblock \textsc{url:}
  \url{http://intelligence.org/files/QuestionsLogicalUncertainty.pdf}.

\bibitem{halpern2011don}
{\sc J.~Y. Halpern} and {\sc R.~Pass},
\newblock {\em arXiv preprint arXiv:1106.2657}  (2011).

\bibitem{halpern2014decision}
{\sc J.~Y. Halpern}, {\sc R.~Pass}, and {\sc L.~Seeman},
\newblock {\em Topics in Cognitive Science} {\bf 6}, 245\textendash257 (2014).

\bibitem{soares2014value}
{\sc N.~Soares},
\newblock The Value Learning Problem,
\newblock Technical report, Machine Intelligence Research Institute, Berkeley,
  2014.

\bibitem{soares2014aligning}
{\sc N.~Soares} and {\sc B.~Fallenstein},
\newblock Aligning Superintelligence with Human Interests: A Technical Research
  Agenda,
\newblock Technical report, Machine Intelligence Research Institute, Berkeley,
  California, 2014.

\bibitem{bostrom2014superintelligence}
{\sc N.~Bostrom},
\newblock {\em Superintelligence: Paths, dangers, strategies},
\newblock Oxford University Press, 2014.

\bibitem{russell1998learning}
{\sc S.~Russell},
\newblock Learning Agents for Uncertain Environments,
\newblock in {\em Proceedings of the Eleventh Annual Conference on
  Computational Learning Theory}, p. 101\textendash103, 1998.

\bibitem{ng2000algorithms}
{\sc A.~Y. Ng} and {\sc S.~Russell},
\newblock Algorithms for Inverse Reinforcement Learning,
\newblock in {\em Proceedings of the 17th International Conference on Machine
  Learning}, p. 663\textendash670, 2000.

\bibitem{chu2005preference}
{\sc W.~Chu} and {\sc Z.~Ghahramani},
\newblock Preference learning with Gaussian processes,
\newblock in {\em Proceedings of the 22nd international conference on Machine
  learning}, p. 137\textendash144, ACM, 2005.

\bibitem{yampolskiy2012leakproofing}
{\sc R.~Yampolskiy},
\newblock {\em Journal of Consciousness Studies} {\bf 19}, 1\textendash2
  (2012).

\bibitem{omohundro2007nature}
{\sc S.~M. Omohundro},
\newblock The nature of self-improving artificial intelligence, 2007,
\newblock Presented at Singularity Summit 2007.

\bibitem{bostrom2012superintelligent}
{\sc N.~Bostrom},
\newblock {\em Minds and Machines} {\bf 22}, 71\textendash85 (2012).

\bibitem{wissner2013causal}
{\sc A.~Wissner-Gross} and {\sc C.~Freer},
\newblock {\em Physical review letters}  (2013),
\newblock 110.16: 168702.

\bibitem{soares2014corrigibility}
{\sc N.~Soares}, {\sc B.~Fallenstein}, {\sc E.~Yudkowsky}, and {\sc
  S.~Armstrong},
\newblock Corrigibility,
\newblock in {\em AAAI-15 Workshop on AI and Ethics}, 2015.

\bibitem{hibbard2012avoiding}
{\sc B.~Hibbard},
\newblock Avoiding unintended AI behaviors,
\newblock in {\em Artificial General Intelligence}, edited by {\sc J.~Bach},
  {\sc B.~Goertzel}, and {\sc M.~Iklé}, p. 107\textendash116, Springer, 2012.

\bibitem{hibbard2014ethical}
{\sc B.~Hibbard},
\newblock {\em Ethical Artificial Intelligence},
\newblock 2014.

\bibitem{hibbard2015self}
{\sc B.~Hibbard},
\newblock Self-Modeling Agents and Reward Generator Corruption,
\newblock in {\em Proceedings of the AAAI-15 Workshop on AI and Ethics}, p.
  61\textendash64, AAAI, 2015.

\bibitem{horvitz2014hundred}
{\sc E.~Horvitz},
\newblock One-Hundred Year Study of Artificial Intelligence: Reflections and
  Framing,
\newblock White paper, Stanford University, 2014.

\bibitem{chalmers2010singularity}
{\sc D.~Chalmers},
\newblock {\em Journal of Consciousness Studies} {\bf 17}, 7\textendash65
  (2010).

\end{thebibliography}

\end{document}